\documentclass[preprint,12pt]{elsarticle}
  \usepackage[en-US]{datetime2}
  \usepackage{amssymb}
  \usepackage{amsmath}
  \usepackage{graphicx}
  \usepackage{url}

\begin{document}

  \begin{frontmatter}

  \title{A.R.I.S.: Automated Recycling Identification System for E-Waste Classification Using Deep Learning}

  \author{Dhruv Talwar, Harsh Desai, Wendong Yin, Goutam Mohanty, Rafael Reveles}

  \address{{\rm Apple}}

\begin{abstract}

Traditional electronic recycling processes suffer from significant resource loss due to inadequate material separation and identification capabilities, limiting material recovery \cite{apple_epr2024}. We present A.R.I.S. (Automated Recycling Identification System), a low-cost, portable sorter for shredded e-waste that addresses this efficiency gap. The system employs a YOLOx model to classify metals, plastics, and circuit boards in real time, achieving low inference latency with high detection accuracy. Experimental evaluation yielded 90\% overall precision, 82.2\% mean average precision (mAP), and 84\% sortation purity. By integrating deep learning with established sorting methods, A.R.I.S. enhances material recovery efficiency and lowers barriers to advanced recycling adoption.
This work complements broader initiatives in extending product life cycles, supporting
trade-in and recycling programs, and reducing environmental impact across the supply chain \cite{apple_epr2024, apple_supply2024}.

\end{abstract}


\end{frontmatter}

\section{Introduction}

Electronic waste (e-waste) is a rapidly growing global issue, with its generation rate far exceeding current recycling efforts \cite{unitar2024}. Each year, tens of millions of tonnes of electronics are discarded, much of which contains valuable and recoverable materials that are not properly recycled. Improper disposal can release substances such as lead, mercury, and brominated flame retardants into the environment, contaminating soil and groundwater and posing risks to ecosystems and human health \cite{aeiscreens}. The volume and complexity of modern electronics also strain existing recycling infrastructures. To address these challenges, Apple collaborates with recyclers to develop and share new disassembly and material recovery solutions, aiming to maximize resource extraction and minimize waste. These efforts align with broader sustainability goals, including trade-in and recycling programs designed to keep products out of landfills \cite{apple_epr2024, apple_supply2024}. In parallel, we are investing in recycling innovations such as those outlined in this paper, with the goal of creating scalable solutions that contribute to industry-wide impact.

\vspace{2mm}

Traditional e-waste recycling processes typically begin with a manual dismantling phase, where components such as batteries are removed. This is followed by mechanical shredding, which shears the remaining components into fragments. The shredded material—a heterogeneous mix of plastics, metals, circuit boards, and glass—is then subjected to various separation techniques. Ferrous metals are extracted using magnetic separators, while non-ferrous metals such as aluminum and copper are isolated via eddy current separators, and optical systems are employed to sort plastics and glass \cite{recyclingtoday}. However, these conventional methods tend to work effectively only for broad material categories. 
They often struggle when tasked with distinguishing between more granular components, such as separating
circuit board fragments from plastic shards or sorting steel from ferrous pieces with copper still attached, which frequently necessitates additional manual sorting.

\vspace{2mm}
In response to these challenges, the formal recycling industry has increasingly turned to sensor-based and intelligent sorting solutions. High-throughput optical sorters, for example, leverage machine vision and near-infrared (NIR) sensors to identify materials based on color or spectral signatures, subsequently employing rapid air-jet actuators to divert targeted items \cite{cpgrp}. Advanced techniques such as X-ray fluorescence (XRF) and inductive metal detection have further refined material identification. Moreover, recent developments have seen the integration of multiple sensor modalities—such as high-resolution cameras, NIR spectroscopy, and electromagnetic detectors—into a single system to enhance sorting accuracy \cite{cpgrp}, \cite{tier1}. Despite their effectiveness, these sophisticated systems often involve high capital and operational costs and require meticulous calibration for each waste stream, rendering them less accessible to smaller recycling operations or regions with limited infrastructure \cite{tier1}.

\vspace{2mm}

Parallel to these sensor-based advancements, significant strides in machine learning, and particularly deep learning, have opened new avenues for addressing e-waste management challenges. Research in this area has demonstrated the potential of advanced neural network architectures to accurately classify and sort shredded e-waste. For instance, sequential neural network (SNN) models have shown promise in classifying distinct e-waste categories \cite{mdpi_snn}, while vision-based detection systems have successfully isolated metal fragments such as copper using robotic implementations \cite{rutgers}. Further studies have applied deep learning to estimate integrated circuit (IC) areas on circuit boards, thereby predicting recoverable metal content \cite{mdpi_ic}. Additionally, high-accuracy object detection models, including YOLOv7, have been utilized to identify materials such as aluminum, copper, circuit boards, plastics, and steel; however, many of these systems have not yet integrated in an actual sortation mechanism \cite{sciencedirect_yolo}. Complementary approaches, such as optical recognition using VGGnet convolutional neural network (CNN) models combined with mechanical sorting enhancements like air nozzle arrays, further illustrate the momentum toward automating e-waste recycling \cite{springer_vgg},\cite{sciencedirect_sorting}.

\vspace{2mm}

In light of these developments, this study presents a novel, low-cost, and fully integrated e-waste sorting system that harnesses deep learning for the automated classification and diversion of shredded materials. Our approach leverages a YOLOx model, interfaced with a programmable logic controller (PLC) and a pneumatic paddle system. This integrated solution is designed to classify e-waste fragments into three primary categories—metal, circuit boards, and plastic— reducing reliance on manual sorting, and increasing separation efficiency. Ultimately, this work aims to bridge the gap between cutting-edge machine learning innovations and practical, scalable industrial recycling applications. A critical challenge in this domain is the prevalence of partially
liberated or composite particles—where materials remain bonded after shredding—which conventional separation techniques struggle to process, often relegating such fragments to lower-value streams or manual re-sorting \cite{partial_liberation}. Our deep learning approach addresses this limitation by enabling granular classification of partially liberated components (e.g., circuit board-plastic composites, copper-steel attachments), allowing the system to sort these complex multi-material particles into targeted fractions based on their dominant material composition rather than discarding them as contaminants.

\section{Dataset Acquisition and Construction}
A proprietary dataset was developed by sourcing shredded e-waste from demanufacturing facilities. Desktop and portable computers were manually disassembled to remove batteries and segregate circuit boards, plastics, and glass before shredding. The shredding process employed a 1\,inch screener to ensure a minimum particle size.

The collected materials were manually sorted into three primary categories:
\begin{itemize}\setlength\itemsep{0em}
  \item \textbf{Metals:} Primarily aluminum from enclosures and stands.
  \item \textbf{Plastics:} Derived from internal components such as casings and speakers.
  \item \textbf{Circuit Boards:} Including a diversity of colors (blue, green, and black) which deviate from the traditional green boards.
\end{itemize}

Using a vibratory feeder, materials were deposited as a monolayer on a 64\,inch wide conveyor belt. Three synchronized red-green-blue (RGB) cameras captured high-resolution images (stitched to a composite image of 5760$\times$1200 pixels).

The final dataset consists of 6,000 images with:
\begin{itemize}\setlength\itemsep{0em}
  \item 5,000 instances of  metals,
  \item 5,500 instances of circuit boards,
  \item 5,000 instances of plastics,
\end{itemize}

To account for the pixels per inch (PPI) required for detection, we utilized images with a resolution of 5760×1200. Given the wide aspect ratio, each image was divided into three segments corresponding to the outputs from individual cameras. These segments were then resized to 640×640, ensuring that the PPI was preserved while still providing sufficient feature detail for accurate model detection. 
This extensive and meticulously curated dataset serves as the foundation of our research, providing robust training and validation for the proposed e-waste sorting system.

\section{Physical System Architecture}
The physical setup for this work involved creating a well-organized and structured environment for image capture, material delivery, controls for material flicking, and software infrastructure. It included several key components:

\subsection{Image Capture Hardware}
 Given the cost constraints within the recycling industry, we prioritized cost-effectiveness while ensuring high-quality results. Key selection criteria included cost, scalability, ease of interface with USB/GigE protocol for edge devices, and high frames per second (FPS) to support very fast throughput—critical for recycling applications. 

The camera selected was a Basler acA1920-155uc with a 12\,mm lens, providing a resolution of 2.3\,MP and up to 155\,FPS. Three RGB cameras were required to cover the entire 64\,inch width of the belt. Mechanical mounting and spacing prevented frame overlap, leading to a unified, homogeneous stitched image for the full belt width. The edge of the first camera’s field-of-view (FOV) served as the starting point for the second camera, and similarly for the third camera; final adjustments were made by adjusting the region of interest (ROI) of each camera’s frame. The cameras were mounted 550\,mm above the conveyor to achieve a granular resolution of 3.5 pixels/mm. The positioning of the camera on top of the belt is shown in Figure~\ref{fig:camera}(a)

Synchronous image acquisition was critical to produce a homogeneous stitched image for the entire belt width. An external trigger signal, generated by a pulse width modulation (PWM) signal generator, provided a +5\,V DC step function at a 50\% duty cycle to all three cameras via their GPIO pins. This PWM signal was generated by the PLC controller, enabling unified control of the vision system, conveyor belt, sorter, and inference engine for the machine learning model.

\subsection{Lighting Setup}
Uniform, diffused lighting was required for optimal image quality. External illumination allowed for a lower exposure rate on the cameras, thereby enabling the highest FPS. A 1.6\,m LED light bar with diffusers was used to prevent hotspots and uniformly illuminate the entire FOV. The lighting bar was positioned at a 30-degree angle from the conveyor belt surface at a height of 150\,mm. Figure~\ref{fig:camera}(b) shows the position of bar lights above the conveyor.

\subsection{Material Delivery}
\begin{figure}[!t]
        \centering

        \begin{minipage}[t]{0.45\textwidth}
          \centering
          \includegraphics[width=0.95\textwidth, height=3.5cm,
  keepaspectratio]{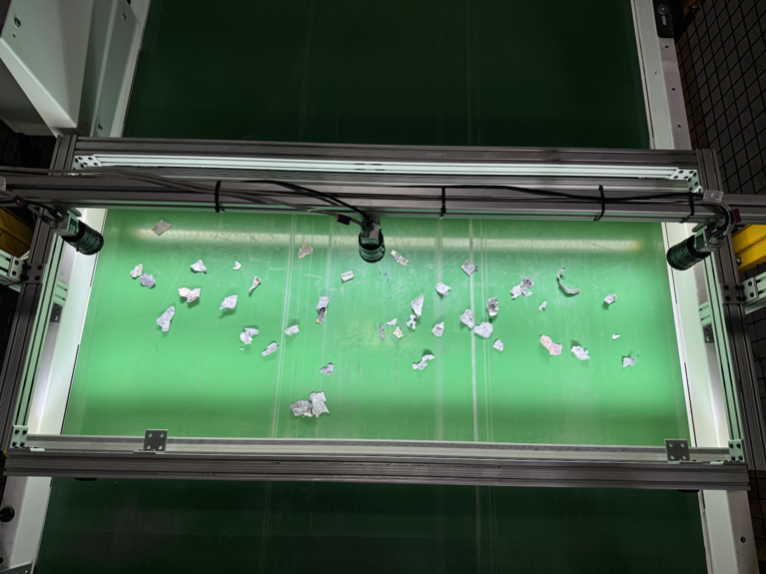}
          \\
          {\small (a) Camera mounting configuration above the conveyor
  belt.}
        \end{minipage}
        \hfill
        \begin{minipage}[t]{0.45\textwidth}
          \centering
          \includegraphics[width=0.95\textwidth, height=3.5cm,
  keepaspectratio]{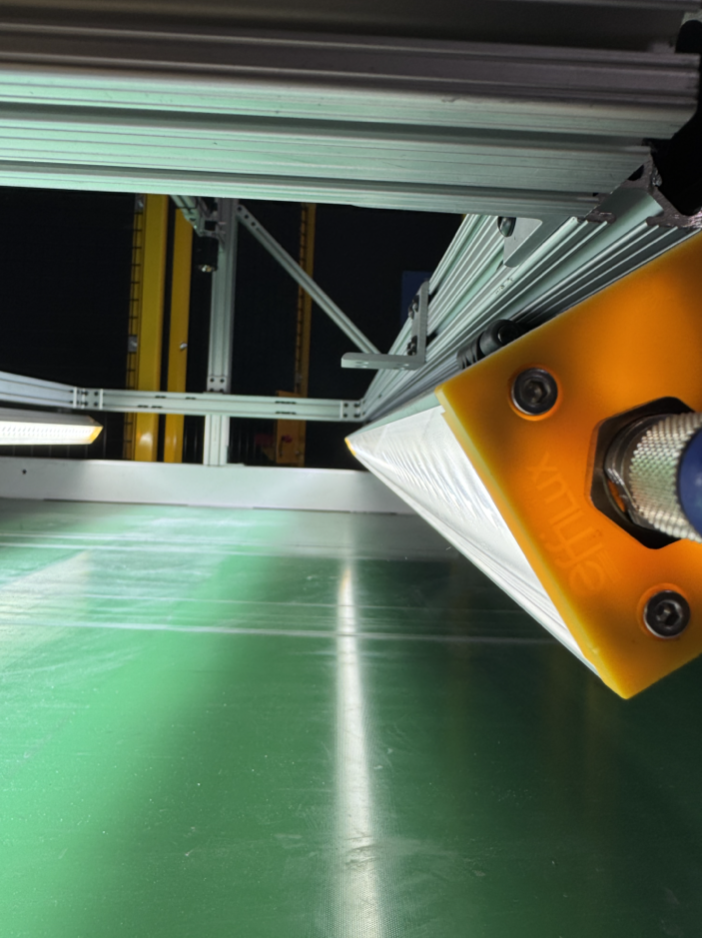}
          \\
          {\small (b) Physical setup for uniform lighting}
        \end{minipage}

        \vspace{0.5cm}

        \begin{minipage}[t]{0.45\textwidth}
          \centering
          \includegraphics[width=0.95\textwidth, height=3.5cm,
  keepaspectratio]{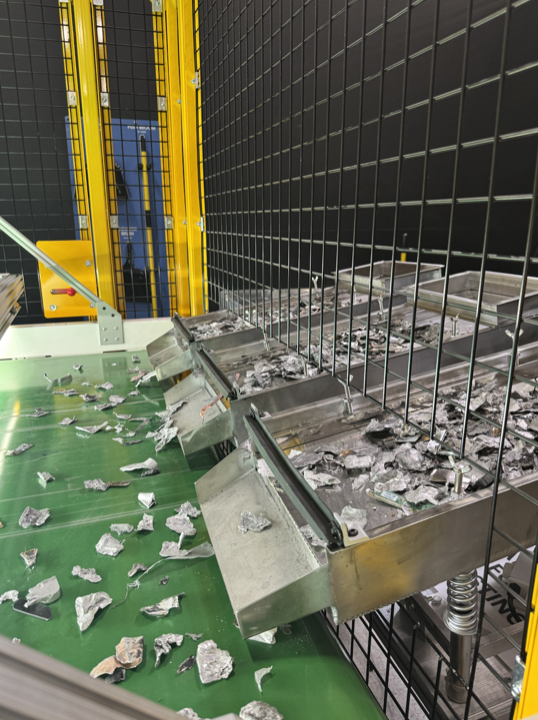}
          \\
          {\small (c) Vibratory feeder distributing shredded e-waste}
        \end{minipage}
        \hfill
        \begin{minipage}[t]{0.45\textwidth}
          \centering
          \includegraphics[width=0.95\textwidth, height=3.5cm,
  keepaspectratio]{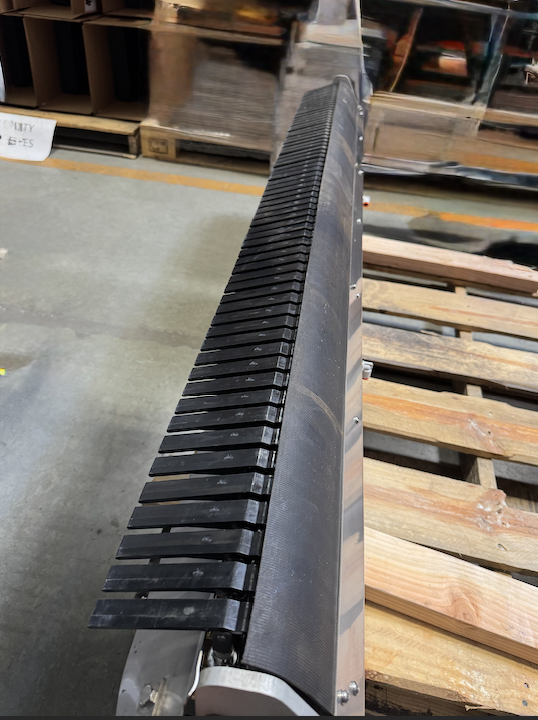}
          \\
          {\small (d) Pneumatic paddle sort device}
        \end{minipage}

        \caption{Physical setup components}
        \label{fig:camera}
  \end{figure}

Surface-based inspection requires that each particle be fully visible to make accurate inferences, as overlapped particles present identification challenges, leading to incorrect classifications. To ensure optimal material presentation, we integrated a vibratory feeder that evenly distributes the shredded e-waste, creating a single-layer deposition on a seamless green conveyor belt measuring 64\,inches in width. The conveyor belt, driven by a variable frequency drive (VFD), maintains a constant speed throughout operation, ranging from 0 to 1.3\,m/s with an operational speed of approximately 1.2\,m/s (245\,ft/min) at a supply frequency of 60\,Hz. This controlled delivery system—combining the vibratory feeder's even distribution with the belt's consistent speed—ensures that each particle is fully exposed to the imaging system while synchronizing image capture and subsequent material sorting, thereby preserving the integrity of the captured data as shown in Figure~\ref{fig:camera}(c).

\subsection{Sortation Mechanism}
We adapted a pneumatic paddle sorter originally used in the agriculture industry for our specific use case shown in Figure~\ref{fig:camera}(d). It consists of 64 paddles, each spaced 1\,inch apart (center-to-center), pneumatically activated by a PLC. Each paddle has a dedicated solenoid wired to the PLC I/O card; applying a 5\,VDC signal to the solenoid triggers actuation. A paddle takes approximately 20\,ms to fully actuate and 20\,ms to return to its home position, resulting in a total cycle time of 40\,ms per flick (25 flicks per second).

Key features of the sorter include:
\begin{itemize}
  \item A pressure regulator that maintains a consistent compressed air supply at 90\,PSI.
  \item A mounting configuration with the sorter positioned 8\,inches from the belt’s edge, and the striking point of each paddle set 6\,inches below the belt’s surface.
\end{itemize}

\subsection{PLC \& Controls}
The system is designed to separate shredded material in real time. It integrates a Siemens S7-1200 PLC and a Mac mini to handle sorting operations. The PLC stores the locations and addresses of each of the 64 paddles and activates the sorter when the corresponding tag is received. By varying the TON (signal ON-time) for each paddle solenoid, we can adjust the actuation time of the paddle.

The S7-1200 PLC also serves as the OPC-UA server, responsible for controlling hardware to actuate up to 64 paddles. It processes incoming data—a string sent by the OPC-UA client (Mac mini)—containing paddle actuation control information (paddle number and flick time, as inferred by the model per frame). A scheduler function ensures precise timing and execution of paddle actuations based on the parameters in each data string. A First In, First Out (FIFO) queue data structure facilitates real-time actuation and de-actuation of paddles, aligning with model inferences for accurate sorting. The PLC program is guided by a finite state machine, optimizing the management of states and events for data reception, parsing, and scheduler control.

\subsection{Computation Hardware}
For real-time decision-making, the edge-based machine learning model is deployed on a Mac mini. This system interfaces with three RGB USB3.0 cameras via Thunderbolt ports and communicates with the PLC over Ethernet using the OPC-UA protocol. This integrated architecture enables rapid processing of incoming frames and timely activation of the appropriate paddles, ensuring seamless operation of the sorting mechanism.

\section{Machine Learning Modeling and Training Subsystem}

\subsection{Object Detection: YoloX}

Our system deploys YOLOx, a state-of-the-art single-stage object detector designed for real-time applications. Unlike traditional two-stage detectors, YOLOx uses an anchor-free approach that directly predicts object locations without predetermined anchor boxes, making it particularly effective for detecting irregularly shaped e-waste fragments \cite{ge2021yolox}. The model features a decoupled head that processes classification and localization tasks separately, improving accuracy while maintaining high inference speed essential for conveyor belt systems.

We initialize YOLOx with weights pre-trained on the COCO dataset, which provides robust feature representations for general object detection. We then fine-tune the model on our custom e-waste dataset, allowing it to specialize in detecting shredded material fragments while retaining the strong feature extraction capabilities learned from COCO. The model's SimOTA assignment strategy dynamically matches predictions to objects based on their quality, handling the challenges of densely packed and overlapping shredded materials effectively.

While YOLOx excels at detecting and localizing e-waste fragments, the bounding boxes it generates are rectangular approximations of irregularly shaped objects. For precise sorting, we further refine these detections to identify the optimal hitting points within each detected fragment, ensuring accurate material separation on the high-speed conveyor belt.

\subsection{Annotation}
Bounding boxes were annotated using the YOLO format: (class, $x_c$, $y_c$, $w$, $h$), where $x_c$ and $y_c$ denote the normalized box center coordinates, $w$, $h$ denote the width and height of the bounding box respectively \cite{ultralytics_yolo_format}. To account for the inherent variability and lack of uniformity in shredded materials, we applied a suite of image augmentation techniques—including light saturation shifts, rotations, and noise injection—to simulate diverse environmental conditions and object appearances. These augmentations are critical for enabling the model to capture and learn intricate features despite the high degree of irregularity present in the dataset \cite{shorten2019}.

The annotation process was carried out in a semi-automated fashion. Initially, approximately 500 images per class were manually annotated. Subsequently, the model was trained on these images—resulting in some degree of overfitting to the specific classes—and then used to annotate subsequent batches. Any missed detections or false negatives were manually corrected. This iterative, semi-automated approach continued until the entire dataset was annotated, allowing us to progressively incorporate and identify corner cases that were underrepresented in the initial dataset. The pipeline for autoannotation is condensed. The continual refinement of the dataset—by identifying and subsequently adding these missing examples—ensured enhanced comprehensiveness and improved model robustness against the diverse manifestations of shredded particles.

\subsection{Training Setup \& Inference Pipeline}

Designing the inference pipeline was a critical step, as it directly governed the trade-off between computational efficiency and real-time performance. A naive strategy of forwarding the full stitched conveyor frame of size $5760 \times 1200$ into the detector would have introduced severe PPI loss due to downscaling and incurred high latency.  

To overcome this, we implemented a batched inference pipeline optimized for throughput. The stitched frame is first partitioned into three segments of size $1920 \times 1200$, each corresponding to a synchronized camera. Every segment is resized to the network input resolution of $640 \times 640$ and assembled into a batch tensor of shape $[3,3,640,640]$. This batch is then forwarded in a single pass through the YOLOx model, enabling parallel inference across all views.  

Preprocessing (color conversion, normalization, resizing) is accelerated using a numba-optimized kernel, while postprocessing employs class-wise non-maximum suppression (NMS) with a confidence threshold of 0.5 and intersection over union (IoU) threshold of 0.5. Detections are remapped to global coordinates in the original $5760 \times 1200$ frame and overlaid with bounding boxes and labels before being passed to the sortation controller.  

This design avoids unnecessary scaling, preserves resolution, and achieves stable real-time performance exceeding 20 FPS on Mac mini with CoreML acceleration. The overall inference pipeline consists of the following steps:

\begin{enumerate}\setlength\itemsep{0em}
    \item Partition stitched frame ($5760 \times 1200$) into 3 segments of $1920 \times 1200$.
    \item Resize each segment to $640 \times 640$ and batch into $[3,3,640,640]$.
    \item Run single batched forward pass through YOLOx model.
    \item Apply class-wise NMS and remap detections to global coordinates.
    \item Render bounding boxes and forward results to the sortation controller.
\end{enumerate}

The model is trained on approximately 6,000 images from three e-waste classes (metals, circuit boards, and plastics), annotated in Yolo format. Data augmentation includes random brightness/contrast adjustments, gamma correction, horizontal flips, rotations, and affine transformations.

Thorough preprocessing was performed before training. Bounding box coordinates were normalized, noisy annotations were filtered out, and degenerate zero-area bounding boxes were removed. At inference time, the optimized model processes each frame in roughly 60\,ms, achieving near real-time performance suitable for deployment on a conveyor-based sorting line.

\section{Material Separation Methodology}
\subsection{Paddle Number}
\begin{figure}[htbp]
    \centerline{\includegraphics[width=0.4\textwidth]{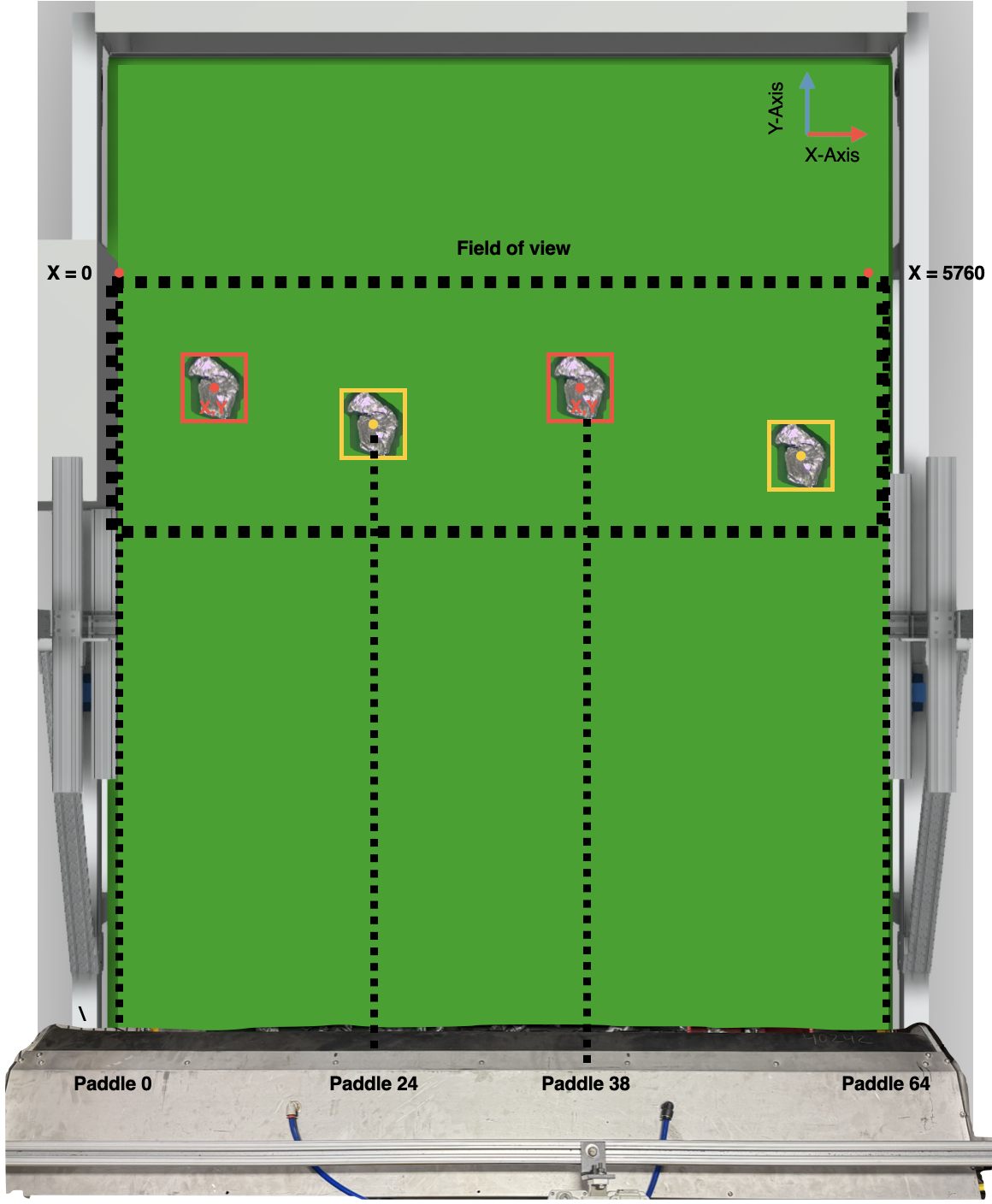}}
    \caption{Physical mapping for paddle number calculation}
    \label{fig:paddle}
\end{figure}

After completing the detection process, the model generates bounding boxes containing each fragment’s class and centroid coordinates. The centroid's $x$-coordinate, which aligns with the belt’s width, is used to map the pixel coordinate to the appropriate paddle number as the fragment approaches. This mapping enables precise paddle selection.  Figure~\ref{fig:paddle} illustrated the mapping.

\subsection{Time of Flicking}
Once the coordinates of the material of interest are detected, the $y$-coordinate is used to ascertain the fragment’s vertical position within the FOV. The conveyor belt length preceding the FOV remains constant, as does the belt’s speed. Consequently, the $y$-coordinate is employed to calculate the total flick time as follows:
\[
T_{\text{flick}} = T_{\text{belt-edge}} + T_{\text{to-hit}} + T_{\text{offset}},
\]
where:
\begin{itemize}
  \item $T_{\text{belt-edge}}$ is the time the particle takes to reach the edge of the belt,
  \item $T_{\text{to-hit}}$ is the (constant) time for the fragment to fall from the belt edge to the optimal paddle hitting point, and
  \item $T_{\text{offset}}$ is a small timing offset determined empirically via real-time testing.
\end{itemize}
By summing these time intervals, we determine the total flick time for each fragment.

\subsection{Material Collection and System Integration}
The sorting system employs a binary collection approach, utilizing two separate bins:
\begin{itemize}
  \item \textbf{Positive Fraction Bin:} Receives particles accurately detected and acted upon by the flicking mechanism.
  \item \textbf{Negative Fraction Bin:} Collects particles that either miss the target or are not classified as relevant, allowing them to continue their original trajectory.
\end{itemize}

\section{System Methodology}
The overall system methodology is as follows:
\begin{enumerate}
  \item The vibratory feeder delivers shredded e-waste (metals, circuit boards, and plastics) onto the conveyor belt in a single layer.
  \item Three cameras capture the material in real time.
  \item The ML model processes the images, calculates the appropriate paddle number based on the refined $x$-coordinate, and determines the flicking time using the $y$-coordinate.
  \item Commands are transmitted via the OPC-UA protocol from the Python-enabled Mac mini to a central PLC, which controls the repurposed sorter as the flicking mechanism.
  \item The system logs all operational data (fragment IDs, packet timestamps, flick execution counts, and any communication breaches) into CSV files for performance analysis.
\end{enumerate}
This integrated architecture enables precise and efficient sorting, ensuring a high concentration of the selected material in the positive fraction while minimizing contamination from other materials.

\section{Experimental Results}

\subsection{Model Training and Evaluation}

\begin{figure}[htbp]
    \centering

    \begin{minipage}[b]{0.48\textwidth}
      \centering
      \includegraphics[width=\textwidth]{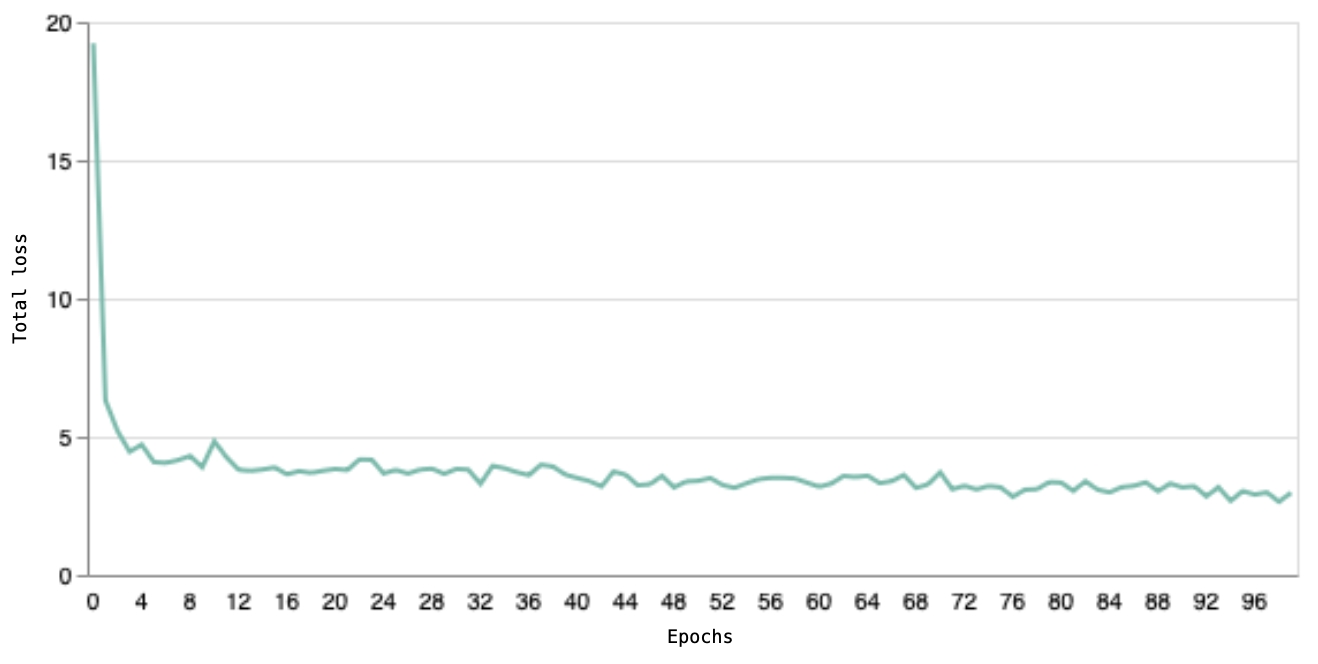}
      \vspace{\fill} 
      \\ 
      (a) Training loss across 100 epochs.
    \end{minipage}
    \hfill
    \begin{minipage}[b]{0.48\textwidth}
      \centering
      \includegraphics[width=\textwidth]{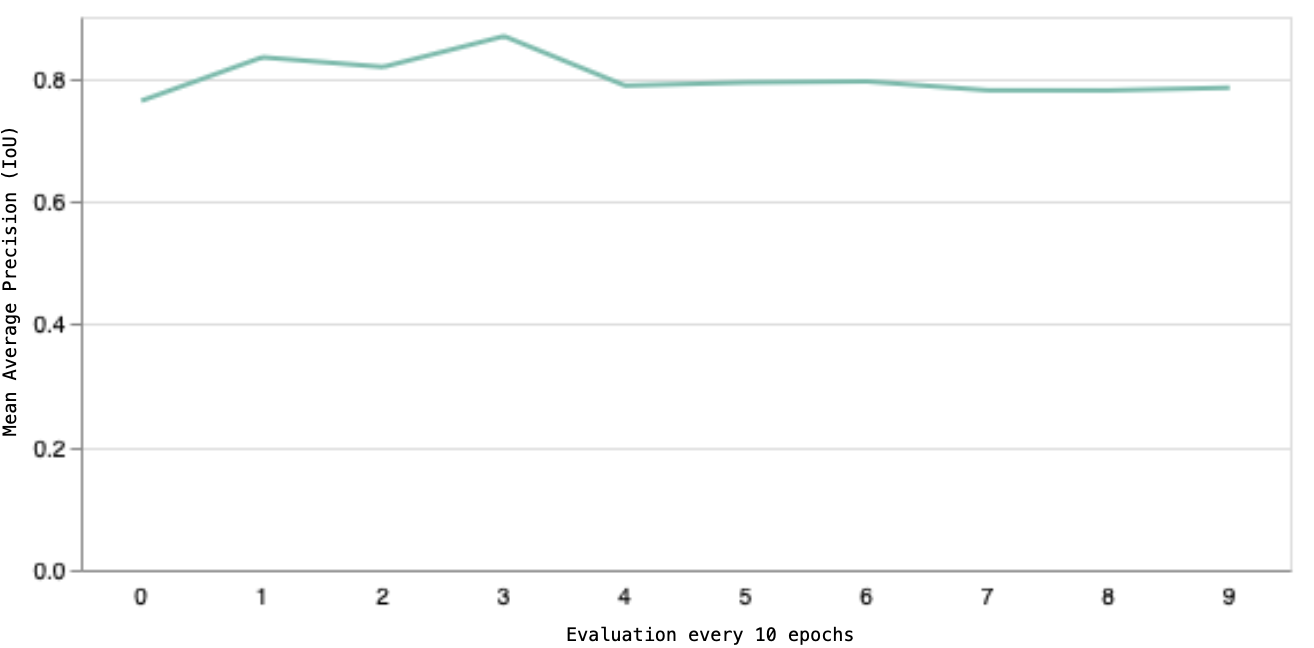}
      \vspace{\fill} 
      \\ 
      (b) mAP@0.50 evaluation.
    \end{minipage}

    \caption{Training \& Evaluation metrics}
    \label{fig:metric_1}
  \end{figure}

The YOLOx model demonstrated strong performance in multi-class e-waste detection and sorting. After 100 training epochs, both training and validation loss curves converged closely, indicating stable learning with minimal overfitting. As shown in Figure~\ref{fig:metric_1}(a), the loss steadily decreased until around epoch 70, after which improvements were marginal.

Evaluation was conducted on a held-out test set of 1,000 images with a batch size of 4. Training employed the Adam optimizer with a cosine learning rate schedule (initial learning rate $1 \times 10^{-2}$, 5-epoch warmup, decaying to 5\% of the maximum value). Data augmentation techniques—including mosaic, mixup, HSV perturbation, horizontal flipping, and affine transformations—further improved generalization.

\begin{figure}[htbp]
    \centering

    \begin{minipage}[b]{0.45\textwidth}
      \centering
      \includegraphics[width=0.5\textwidth]{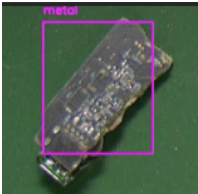}
      \vspace{\fill}
      \\ 
      {\small (a) Incorrect detection}
    \end{minipage}
    \hfill
    \begin{minipage}[b]{0.45\textwidth}
      \centering
      \includegraphics[width=0.5\textwidth]{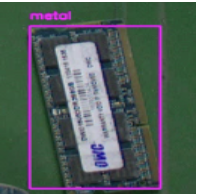}
      \vspace{\fill}
      \\ 
      {\small (b) Bright white surface}
    \end{minipage}

    \vspace{0.3cm} 

    \begin{minipage}[b]{0.45\textwidth}
      \centering
      \includegraphics[width=0.5\textwidth]{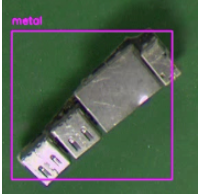}
      \vspace{\fill}
      \\ 
      {\small (c) USB port attached }
    \end{minipage}
    \hfill
    \begin{minipage}[b]{0.45\textwidth}
      \centering
      \includegraphics[width=0.5\textwidth]{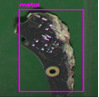}
      \vspace{\fill}
      \\ 
      {\small (d) Exposed shiny surface}
    \end{minipage}

    \vspace{0.5cm} 

    \begin{minipage}[b]{0.45\textwidth}
      \centering
      \includegraphics[width=\textwidth]{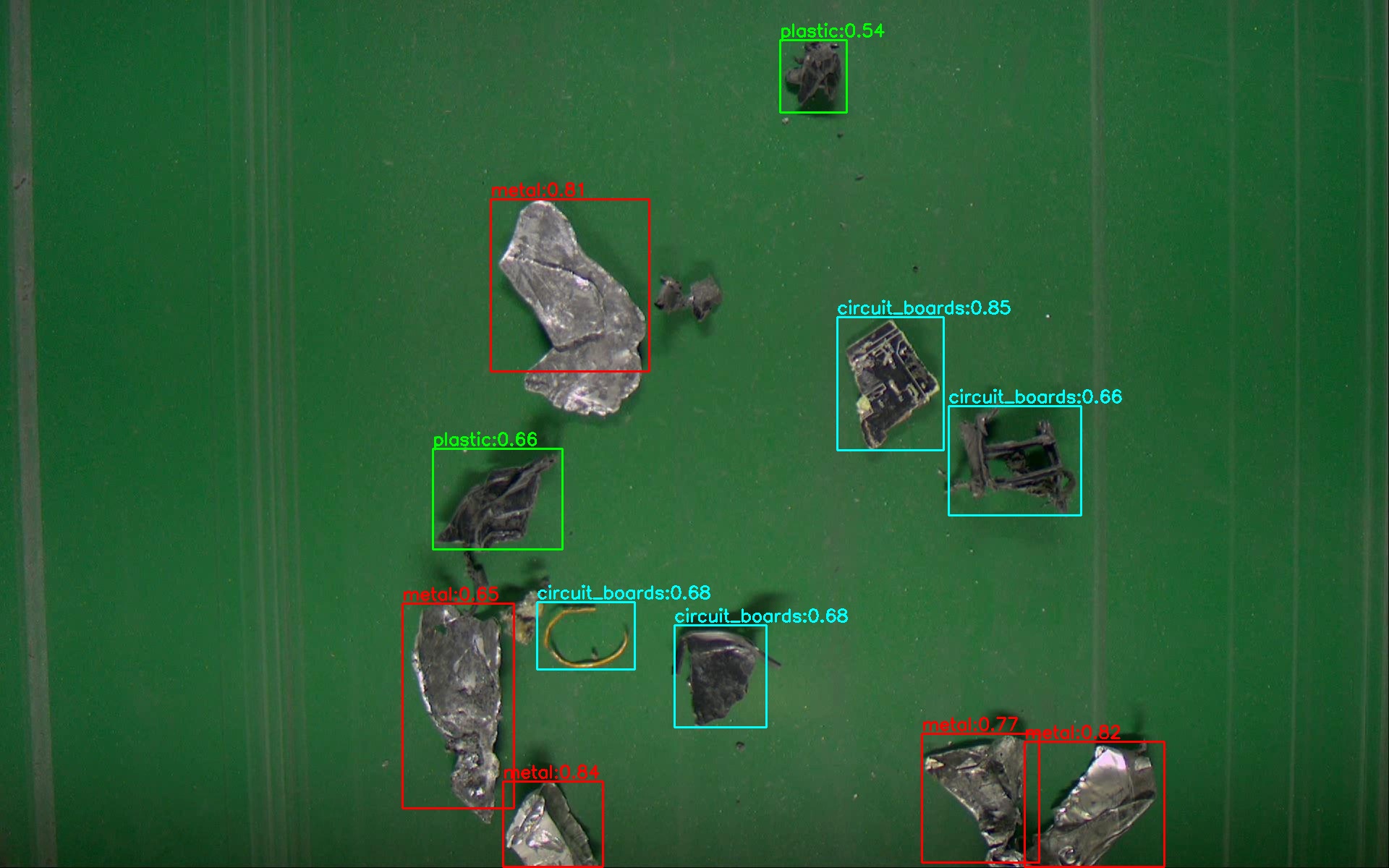}
      \vspace{\fill}
      \\ 
      {\small (e) Detection frame 1}
    \end{minipage}
    \hfill
    \begin{minipage}[b]{0.45\textwidth}
      \centering
      \includegraphics[width=\textwidth]{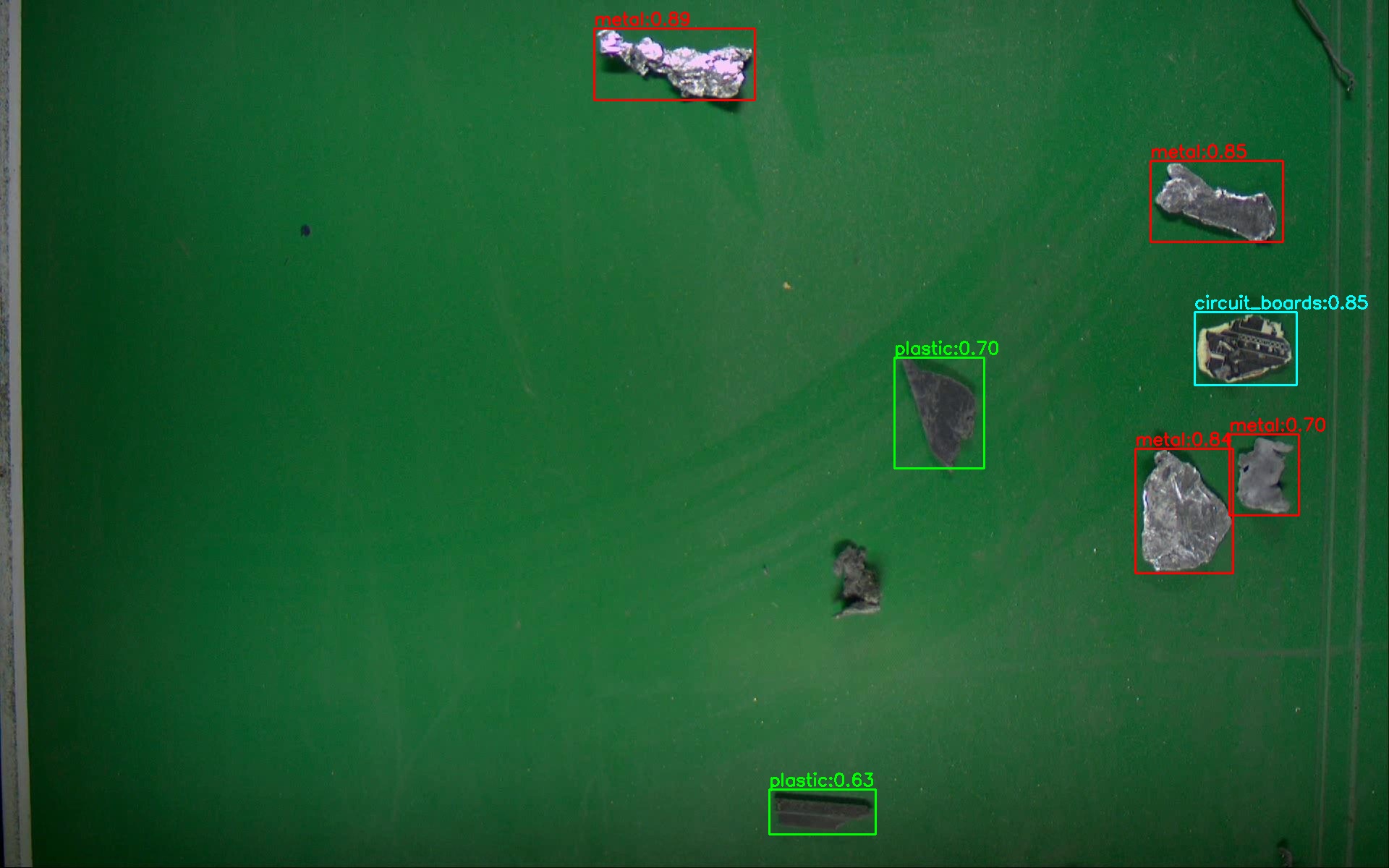}
      \vspace{\fill}
      \\ 
      {\small (f) Detection frame 2}
    \end{minipage}

    \caption{Edge cases and sample detections: (a-d) corner cases, (e-f) successful detections}
    \label{fig:detections}
\end{figure}

Several challenging cases were observed during training. Circuit boards adhered to metallic fragments, contributed to class imbalance, with the detector biased toward metals (Figure~\ref{fig:detections}(a)). Lighting variations caused bright white regions on boards to be misclassified as metal, shown in Figure~\ref{fig:detections}(b), while USB ports and transistors with metallic features in Figure~\ref{fig:detections}(c) and exposed shiny surfaces Figure~\ref{fig:detections}(d) also produced false positives. In other cases, boards with plastic-like textures were confused with plastics. These corner cases Figure~\ref{fig:detections}(a-d) highlight the classification challenges inherent in heterogeneous e-waste streams.

To mitigate these issues, the dataset was augmented with additional corner-case images, while enhanced augmentation techniques were applied to simulate lighting variations and color distortions. The augmented dataset included circuit boards with high metal concentration, visually plastic-like boards, and boards with shiny white surfaces. Transfer learning from pre-trained weights further improved performance metrics for circuit boards and plastics. Table~\ref{tab:classwise} summarizes the resulting class-wise performance.

\subsubsection{Class-wise Performance}

The refined dataset and training strategy yielded strong detection performance across all material classes. Table~\ref{tab:classwise} presents the precision, recall, and average precision (AP) values for each class. Metals achieved the highest AP of 88.9\%, while circuit boards reached 87.2\%. Plastics achieved very high precision 99.7\% but lower recall 56.2\%, indicating that the model is conservative when predicting plastics—false positives are rare, but many true instances are missed. Figure~\ref{fig:detections}(e-f) illustrates real-world detection outputs from the trained model. These frames are shown to demonstrate actual detections on e-waste samples.

\begin{table}[htbp]
\caption{Class-wise performance metrics.}
\label{tab:classwise}
\centering
\begin{tabular}{lccc}
\hline
\textbf{Class} & \textbf{Precision (\%)} & \textbf{Recall (\%)} & \textbf{AP (\%)} \\
\hline
Metals         & 92.8 & 86.3 & 88.9 \\
Circuit Boards & 78.5 & 94.1 & 87.2 \\
Plastics       & 99.7 & 56.2 & 70.4 \\
\hline
\end{tabular}
\end{table}

\subsubsection{Precision--Recall Analysis}

Precision--Recall (PR) curves Figure~\ref{fig:performance_metrics}(a) provide insight into class-specific detection behavior across varying thresholds. The model achieved AP values of 87.9\% for circuit boards, 88.9\% for metals, and 70.4\% for plastics, yielding an overall mAP of 82.2\%.

For comparison, random classifier baselines were estimated from dataset priors. The distribution of material classes in the test dataset is as follows:
\begin{itemize}\setlength\itemsep{0em}
  \item \textbf{Circuit Boards:} 729 samples (39.0\%)
  \item \textbf{Metals:} 534 samples (28.5\%)
  \item \textbf{Plastics:} 608 samples (32.5\%)
  \item \textbf{Total:} 1,871 samples (100\%)
\end{itemize}
Since a random predictor assigns labels according to class frequency, the expected AP values are 39\% for circuit boards, 28.5\% for metals, and 32.5\% for plastics. The trained model substantially outperforms these baselines, confirming that it learned meaningful distinguishing features. Notably, while plastics remain the most challenging class, their AP of 70.4\% is still more than double the random baseline. Figure~\ref{fig:performance_metrics}(a) compares PR curves across all three classes, with metals achieving the highest area under the curve (AUC), followed by circuit boards, then plastics.

\subsubsection{Confusion Matrix Analysis}

The normalized confusion matrix in Figure~\ref{fig:performance_metrics}(b) highlights class-specific challenges. Circuit boards and metals achieved high recall values of 94.1\% and 86.3\%, respectively. Plastics, however, showed low recall of 56.2\% despite near-perfect precision.
The combination of low recall and near-perfect precision for plastics indicates that the model is highly conservative when labeling plastics, minimizing false positives but missing many true positives.

\subsubsection{Overall Detection Performance}

Detection quality was evaluated using mean average precision (mAP) at different Intersection-over-Union (IoU) thresholds. The mAP evaluation was done at every 10th iteration. On the validation set, the model achieved an mAP@0.50 of 78.5\% and a stricter COCO-style mAP@0.50:0.95 of 52.7\%, as shown in Figure~\ref{fig:metric_1}(b). These values reflect stable convergence and demonstrate that the model is generalized effectively without overfitting.

\begin{figure}[t]
    \centering

    \includegraphics[width=1\textwidth]{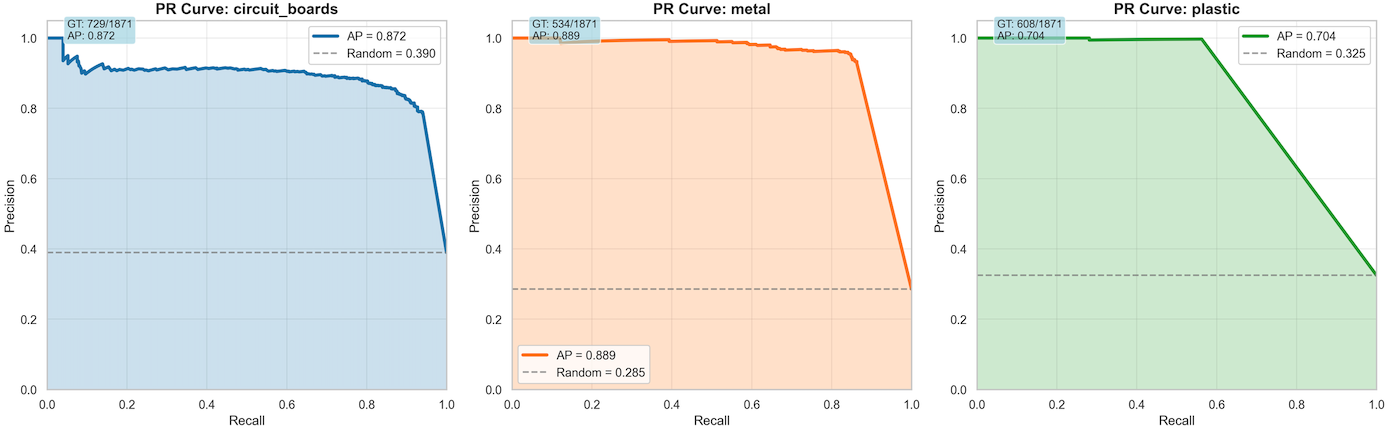} 
    \vspace{0.2cm} 
    {\small (a) Precision–Recall curves for circuit boards, metals, and plastics.}

    \begin{minipage}[b]{0.3\textwidth}  
      \centering
      \includegraphics[width=\textwidth]{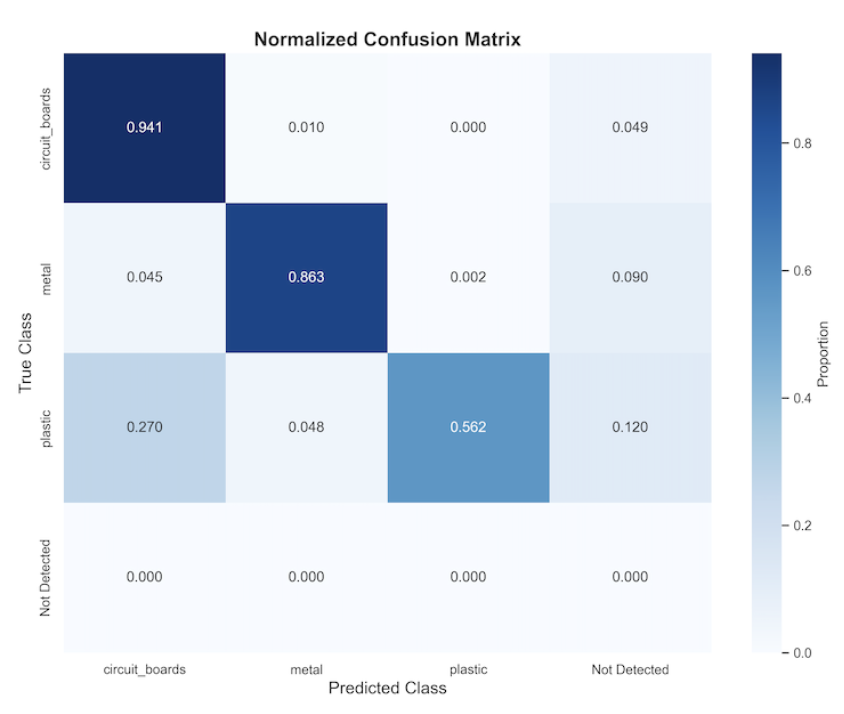}
      \vspace{\fill}
      \\
      {\small (b) Normalized confusion matrix across classes.}
    \end{minipage}
    \hfill
    \begin{minipage}[b]{0.3\textwidth}
      \centering
      \includegraphics[width=\textwidth]{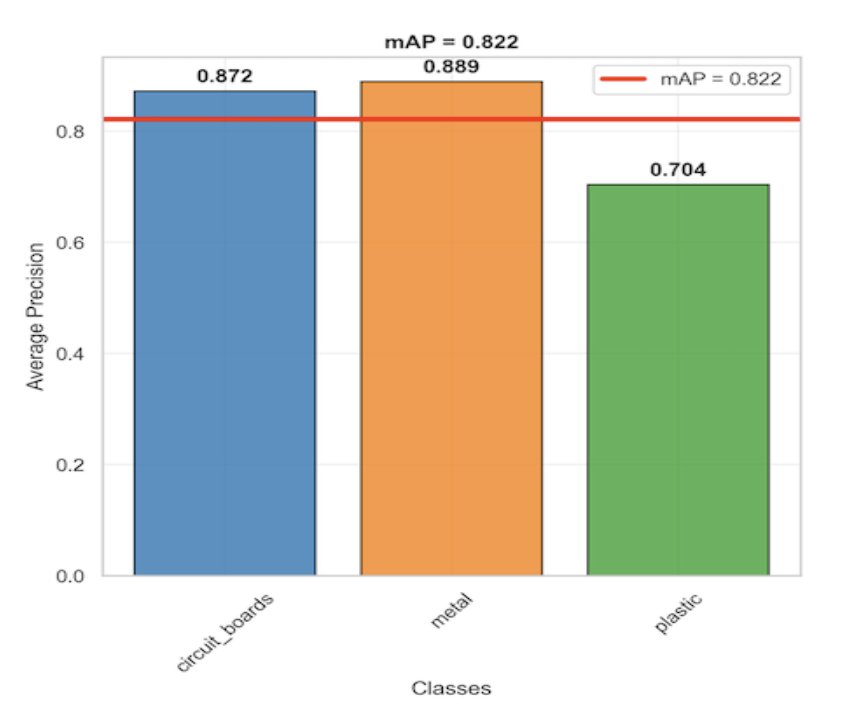}
      \vspace{\fill}
      \\
      {\small (c) Average Class Precision and Overall mAP.}
    \end{minipage}
    \hfill
    \begin{minipage}[b]{0.3\textwidth}
      \centering
      \includegraphics[width=\textwidth]{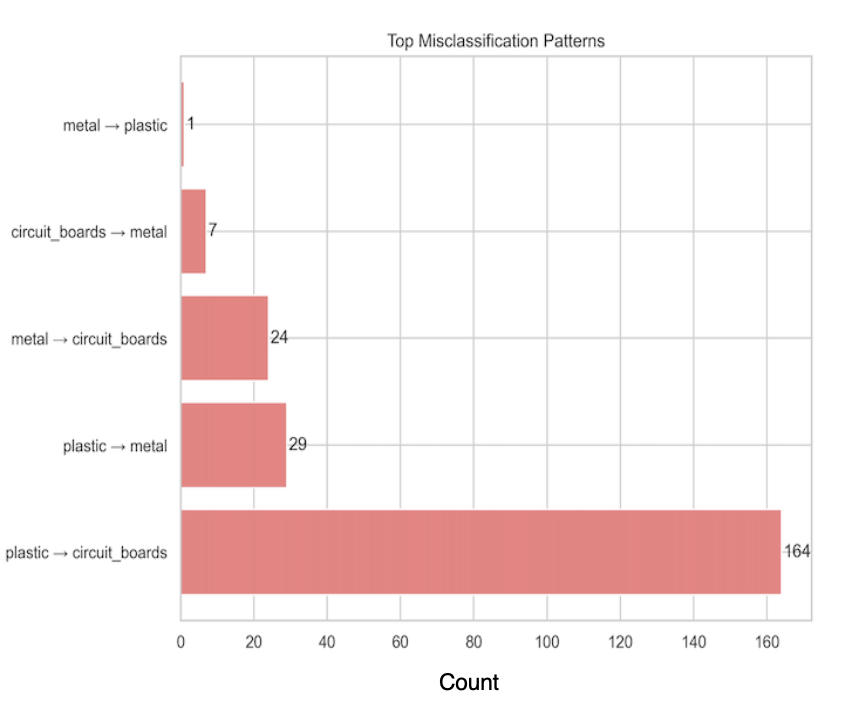}
      \vspace{\fill}
      \\
      {\small (d) Frequency of misclassification patterns.}
    \end{minipage}

    \vspace{0.3cm}  
    \caption{Model performance evaluation metrics and classification analysis}
    \label{fig:performance_metrics}
  \end{figure}

On the held-out test set, the model achieved an mAP@0.50 of 82.2\%. The result the model achieved on the test set closely matches the validation performance 78.5\%, with only a small difference of 3.7\%. This indicates consistent generalization to unseen data, and thus the reported test performance can be considered a reliable indicator of real-world behavior. Figure~\ref{fig:performance_metrics}(c) shows the average precision per class along with the overall mAP.

\subsubsection{Dataset Composition and Misclassification Patterns}
The test dataset contained a balanced distribution of material
types. Misclassification analysis Figure~\ref{fig:performance_metrics}(d) revealed
that plastics were the most frequently confused class: 164
out of 608 (27.0\%) plastic samples were mislabeled as circuit boards, and 20 out of 608 (4.8\%) were mislabeled as metals. This
confusion primarily arises from visual similarities between
plastic components with embedded circuitry and actual circuit
boards, particularly under reflective lighting conditions.

\subsection{Physical Sortation Results}
Once the model demonstrated stable performance on the test set --- with an mAP@0.50 of 82.2\% and class-wise recall values of 86.3\% for metals, 94.1\% for circuit boards, and 56.2\% for plastics --- it was deployed to evaluate physical sorting performance in a controlled setting. A 100\,lb batch of mixed electronic waste (metals, circuit boards, and plastics) was conveyed at 1.3\,m/s through the sorter, and the recovery purity of each target material stream was measured.

Detection performance remains stable up to an IoU of about 0.7, after which the detection rate drops sharply, indicating that the model produces very tight bounding boxes around objects, which ensures accurate centroid localization. Accurate centroids are critical in this application, since they directly determine paddle actuation timing and hence sortation accuracy.

The recovery results for the sorter are summarized below. When configured to target a single material per run, the system achieved:

\begin{itemize}\setlength\itemsep{0em}
  \item \textbf{Metals:} 89\% purity (Metal vs. Other)
  \item \textbf{Circuit Boards:} 85\% purity (Circuit board vs. Other)
  \item \textbf{Plastics:} 79\% purity (Plastic vs. Other)
\end{itemize}

The slightly lower purity for plastics reflects the model’s conservative recall for that class, as also observed in the detection metrics.

In terms of throughput, the system achieved an average sortation rate of approximately 5\,kg/s under standard operating conditions. Some operational challenges were observed due to irregular particle shapes and random orientations, occasionally leading to paddle strikes at non-optimal points and thus imperfect trajectories. These errors are currently being mitigated through improved centroid tracking and refined actuation timing modules.

\subsection{Practical Implications and Outlook}
These results show that the system can reliably recover nearly all valuable metals and circuit boards in real e-waste streams. High recall for metals (86.3\%) and circuit boards (94.1\%) ensures that most high-value materials are successfully diverted to the correct stream. Although plastic recall remains lower (56.2\%), the very high precision (99.7\%) means that contamination of valuable fractions is minimal. This trade-off is operationally acceptable: missing some plastics has a smaller economic impact than mistakenly misclassifying metals or circuit boards, which would reduce the value of the recovered output.  

From an operational standpoint, the system is readily deployable in real-world settings.  The detector runs at over 20 FPS, making it suitable for real-time conveyor-belt sorting while reducing manual labor and requiring only minimal quality-control checks to handle residual errors. In practical terms, this demonstrates that the system is not just detecting objects accurately in controlled evaluation, but is also achieving meaningful material recovery performance in practice. Metals and circuit boards — the two most valuable components in e-waste streams — are recovered with high recall and precision, substantially improving recycling efficiency.  

Nonetheless, limitations remain. The detector struggles with very small fragments (e.g., plastic crumbs, metallic dust), which are often missed and require pre-sieving. Addressing this will require expanding the dataset with more fine-grained samples, refining augmentation strategies to handle extreme fragmentations, and further optimizing the model for faster inference while improving plastic recall.  

In summary, the proposed system represents a practical step toward scalable, automated e-waste sorting, where high-value recovery can be achieved with minimal human intervention. With targeted improvements, particularly for plastics and small fragments, it has the potential to operate as a cost-effective, industrial-grade solution for sustainable recycling operations.

\vspace{5pt}

\section{Conclusion}
This work presents a cost-effective and scalable e-waste sorting system powered by the YOLOx detector and affordable actuation hardware. The model achieved an mAP of 82.2\% at IoU at 0.50 on the test set, closely matching the validation result of 78.5\%, confirming strong generalization. The IoU–detection rate analysis further showed that bounding boxes remain tight, enabling accurate centroid estimation for reliable paddle strikes.  

In controlled sortation trials, the system achieved material recovery rates of 89\% for metals, 85\% for circuit boards, and 79\% for plastics. In addition, precision–recall analysis revealed high recall for metals (86.3\%) and circuit boards (94.1\%), ensuring that nearly all valuable fractions are recovered, while plastics achieved very high precision (99.7\%), minimizing contamination of higher-value outputs. These results demonstrate that the system can substantially improve recovery of valuable e-waste components, operating at around 20 FPS for real-time deployment on conveyor belts.  

Beyond performance metrics, the system highlights a practical balance between economic and operational considerations: occasional misses in plastics pose limited cost impact, while reliable recovery of metals and circuit boards significantly enhances downstream recycling efficiency. Limitations remain in detecting very small fragments (e.g., plastic crumbs, metallic dust), which future work will address through expanded datasets, fine-grained augmentations, and multi-scale inference strategies.  

In conclusion, this work demonstrates not only high detection accuracy in controlled experiments but also effective translation into real-world material recovery. The integration of iterative data augmentation, model-in-the-loop refinement, and cost-efficient hardware offers a scalable pathway toward industrial-grade, automated e-waste sorting. With continued improvements targeting plastics and small fragments, the system has strong potential as a globally deployable, sustainable recycling solution.


\begin{thebibliography}{99}\setlength\itemsep{0em}

\bibitem{apple_epr2024}
Apple, ``Environmental Progress Report, 2024,'' Cupertino, CA, USA: Apple Inc., Apr. 2025.

\bibitem{apple_supply2024}
Apple, ``People and Environment in Our Supply Chain Report, 2024,'' Cupertino, CA, USA: Apple Inc., 2024.

\bibitem{unitar2024}
UNITAR, ``Global E-Waste Monitor 2024: Electronic Waste Rising Five Times Faster than Documented E-Waste Recycling,'' 2024. [Online]. Available: https://www.unitar.org/about/news-stories/news/global-e-waste-monitor-2024-electronic-waste-rising-five-times-faster-documented-e-waste-recycling

\bibitem{aeiscreens}
ITAD USA, ``The Hidden Dangers of Improper E-Waste Disposal,'' 2025. [Online].

\bibitem{recyclingtoday}
Recycling Today, ``Modern E-Waste Recycling Processes,'' [Online].

\bibitem{cpgrp}
CPGRP, ``Advanced E-Waste Sorting Solutions,'' [Online].

\bibitem{tier1}
Tier1, ``From Waste to Worth: The Circular ESG Initiative Driving Sustainable Change,'' [Online].

\bibitem{mdpi_snn}
G. P. Oise and S. Konyeha, ``Deep learning system for e-waste management,'' \textit{Engineering Proceedings}, vol. 67, no. 1, p. 66, 2024.

\bibitem{rutgers}
S. D. Han \textit{et al.}, ``Toward fully automated metal recycling using computer vision and non-prehensile manipulation,'' in \textit{Proc. 2021 IEEE 17th Int. Conf. Automation Science and Engineering (CASE)}, Aug. 2021, pp. 891--898.

\bibitem{mdpi_ic}
L. H. D. S. Silva \textit{et al.}, ``Estimating recycling return of integrated circuits using computer vision on printed circuit boards,'' \textit{Applied Sciences}, vol. 11, no. 6, p. 2808, 2021.

\bibitem{sciencedirect_yolo}
P. K. Sarswat, R. S. Singh, and S. V. S. H. Pathapati, ``Real-time electronic-waste classification algorithms using computer vision based on convolutional neural networks (CNN),'' \textit{Resources, Conservation and Recycling}, vol. 207, p. 107651, 2024.

\bibitem{springer_vgg}
Z. Gao, S. Sridhar, D. E. Spiller, and P. R. Taylor, ``Applying improved optical recognition with machine learning on sorting Cu impurities in steel scrap,'' \textit{Journal of Sustainable Metallurgy}, vol. 6, pp. 785--795, 2020.

\bibitem{sciencedirect_sorting}
S. P. Gundupalli, S. Hait, and A. Thakur, ``A review on automated sorting of source-separated municipal solid waste for recycling,'' \textit{Waste Management}, vol. 60, pp. 56--74, 2017.

\bibitem{partial_liberation}
M. A. Reuter and A. van Schaik, ``Opportunities and limits of recycling: A dynamic-model-based analysis,'' \textit{MRS Bulletin}, vol. 37, no. 4, pp. 339--347, 2012.

\bibitem{ge2021yolox}
Z. Ge, S. Liu, F. Wang, Z. Li, and J. Sun, ``YOLOX: Exceeding YOLO series in 2021,'' arXiv:2107.08430, 2021.

\bibitem{ultralytics_yolo_format}
Ultralytics, ``Object Detection Dataset Formats,'' \textit{Ultralytics YOLO Documentation}, 2025.

\bibitem{shorten2019}
C. Shorten and T. M. Khoshgoftaar, ``A survey on image data augmentation for deep learning,'' \textit{Journal of Big Data}, vol. 6, no. 1, pp. 1--48, 2019.
\end{thebibliography}
\end{document}